\documentclass{article}

\usepackage{arxiv}

\usepackage[utf8]{inputenc} 
\usepackage[T1]{fontenc}    
\usepackage{hyperref}       
\usepackage{url}            
\usepackage{booktabs}       
\usepackage{amsfonts}       
\usepackage{nicefrac}       
\usepackage{microtype}      
\usepackage{lipsum}%
\usepackage{graphicx}%
\usepackage{graphicx}%
\usepackage{multirow}%
\usepackage{amsmath,amssymb,amsfonts}%
\usepackage{amsthm}%
\usepackage[mathscr]{eucal}%
\usepackage[title]{appendix}%
\usepackage{xcolor}%
\usepackage{textcomp}%
\usepackage{manyfoot}%
\usepackage{booktabs}%
\usepackage{algorithm}%
\usepackage{algorithmicx}%
\usepackage{algpseudocode}%
\usepackage{listings}%
\usepackage[edges]{forest}%
\usepackage{pgfplots}%
\pgfplotsset{compat=1.17}%
\usepackage{tikz}%
\usetikzlibrary{arrows.meta}%
\usepackage{array}%
\usepackage{lscape}%
\usepackage{booktabs}%
\usepackage{tabularx}%
\usepackage{rotating}%
\usepackage[utf8]{inputenc}%
\usepackage[T1]{fontenc}%
\usepackage{mathrsfs}%
\usepackage{lmodern}%
\usepackage{orcidlink}%
\usepackage{tikz}%
\usepackage{graphicx}
\usepackage{array}
\usetikzlibrary{shadows, arrows.meta, positioning}%
\graphicspath{ {./images/} }%

\newcommand{\bmsection}[1]{\section*{\textbf{#1}}}

\title{Analysis of Hyperparameter Optimization Effects on Lightweight Deep Models for Real-Time Image Classification}

\author{
Vineet Kumar Rakesh~\orcidlink{0009-0000-7102-6564} \\
Engineering Sciences, Homi Bhabha National Institute \\
Training School Complex, Anushaktinagar, Mumbai, Maharashtra 400094, India \\
Computer and Informatics Group, Variable Energy Cyclotron Centre \\
1/AF, Bidhannagar, Kolkata, West Bengal 700064, India \\
\texttt{vineet@vecc.gov.in} \\
\And
Soumya Mazumdar~\orcidlink{0009-0006-3521-9557} \\
Department of Computer Science and Business Systems \\
Gargi Memorial Institute of Technology \\
Baruipur, Kolkata, West Bengal 700144, India \\
\texttt{reachme@soumyamazumdar.com} \\
\And
Tapas Samanta~\orcidlink{0000-0003-0521-0747} \\
Computer and Informatics Group, Variable Energy Cyclotron Centre \\
1/AF, Bidhannagar, Kolkata, West Bengal 700064, India \\
Engineering Sciences, Homi Bhabha National Institute \\
Training School Complex, Anushaktinagar, Mumbai, Maharashtra 400094, India \\
\texttt{tsamanta@vecc.gov.in} \\
\And
Hemendra Kumar Pandey~\orcidlink{0000-0001-7203-2990} \\
Radioactive Ion Beam Facilities Group, Variable Energy Cyclotron Centre \\
1/AF, Bidhannagar, Kolkata, West Bengal 700064, India \\
Engineering Sciences, Homi Bhabha National Institute \\
Training School Complex, Anushaktinagar, Mumbai, Maharashtra 400094, India\\
\texttt{hemendra@vecc.gov.in} \\
\And
Amitabha Das~\orcidlink{0009-0003-1460-8308} \\
School of Nuclear Studies and Application \\
Jadavpur University \\
Salt Lake City, Kolkata, West Bengal 700106, India \\
\texttt{amitabhad.snsa@jadavpuruniversity.in} \\
}

\begin{document}
\maketitle
\newpage
\begin{abstract}
Lightweight convolutional and transformer-based networks are increasingly preferred for real-time image classification, especially on resource-constrained devices. This study evaluates the impact of hyperparameter optimization on the accuracy and deployment feasibility of seven modern lightweight architectures--ConvNeXt--T, EfficientNetV2 --S, MobileNetV3--L, MobileViT v2 (S/XS), RepVGG--A2, and TinyViT--21M—trained on a class-balanced subset of 90,000 images from ImageNet--1K. Under standardized training settings, this paper investigates the influence of learning rate schedules, augmentation, optimizers, and initialization on model performance. Inference benchmarks are performed using an NVIDIA L40s GPU with batch sizes ranging from 1 to 512, capturing latency and throughput in real--time conditions. This work demonstrates that controlled hyperparameter variation significantly alters convergence dynamics in lightweight CNN and transformer backbones, providing insight into stability regions and deployment feasibility in edge  Artificial Intelligence. Our results reveal that tuning alone leads top-1 accuracy improvement of 1.5–-3.5\% over baselines, and select models (e.g., RepVGG--A2, MobileNetV3--L) deliver sub--5~ms latency and over 9,800 FPS, making them ideal for edge deployment. This work provides reproducible, subset-based insights into lightweight hyperparameter tuning and its role in balancing speed and accuracy. The code and logs may be seen at:~\url{https://vineetkumarrakesh.github.io/lcnn-opt}.
\end{abstract}

\keywords{Lightweight Deep Learning \and Hyperparameter Optimization \and Real-Time Classification \and Accuracy \and Model Efficiency \and AutoML}

\subsection*{Introduction}\label{sec1}

Real-time picture categorization on peripheral devices entails the deployment of deep learning models that not only assure high accuracy but also remain computationally efficient. To resolve this balance, recent research has focused on lightweight architectures, typically under 30 million parameters, including convolutional and hybrid CNN-transformer models. In this work, the paper systematically explores seven such models which includes EfficientNetV2-S~\cite{tan2021efficientnetv2}, ConvNeXt-T~\cite{liu2022convnext}, MobileViT v2~\cite{mehta2023mobilevitv2}, MobileNetV3-L~\cite{howard2019}, TinyViT-21M~\cite{tinyvit2022}, and RepVGG-A2~\cite{ding2021repvgg} selected for their ubiquity and architectural variety. This paper analyzes how intentional hyperparameter adjustment may significantly enhance the accuracy and training behavior of these lightweight models. This study included training and evaluating all models using a class-balanced sample of 90,000 images from ImageNet--1K~\cite{ILSVRC15}. This design decision enables rapid, regulated experimentation while preserving class variety, while absolute accuracy metrics are not directly comparable to comprehensive dataset benchmarks~\cite{mehta2021mobilevit, tinyvit2022}. 

The central hypothesis of this study is that systematic hyperparameter optimization yields statistically significant accuracy and latency improvements in lightweight models without architectural modification. One of the most remarkable discoveries is the significance of the initial learning rate in impacting ultimate model correctness. While previous frameworks automate search via Bayesian or bandit methods, this study isolates and quantifies individual parameter sensitivity, bridging empirical engineering and theoretical optimization understanding. For instance, increasing the learning rate from 0.001 to 0.1 leads to large gains: ConvNeXt-T increases from 77.61\% to 81.61\% Top-1 accuracy, while TinyViT-21M climbs from 85.49\% to 89.49\%. However, further elevating the learning rate to 0.2 produces a reduction in performance, demonstrating a distinct optimal range for convergence stability. Data augmentation and regularization approaches also play a crucial role in enhancing generalization. Starting with a simple training pipeline, the inclusion of RandAugment, Mixup, CutMix~\cite{yun2019cutmix}, and Label Smoothing delivers continuous gains~\cite{ding2021repvgg, tan2019efficientnet}. For example, MobileViT v2 (S) rises from 85.45\% in the baseline configuration to 89.45\% when all augmentation strategies are implemented. A similar pattern is found across all other models, emphasizing the benefits of a composite augmentation pipeline ~\cite{loshchilov2017sgdr, ding2021repvgg}. Training all models with a batch size of 512 not only exploits the memory capacities of high-end GPUs like the NVIDIA L40s but also assures consistent convergence with efficient throughput. While prior works automate HPO via Bayesian search, few isolate the effect of individual hyperparameters across multiple architectures under identical constraints. Convergence trends imply that larger models such as TinyViT-21M and EfficientNetV2-S converge to optimal accuracy more quickly, particularly under cosine learning rate scheduling. The choice of optimizer further influences training dynamics. While SGD with momentum works well across all CNN-based models, transformer-based and hybrid models such as ConvNeXt-T and TinyViT-21M display greater early-stage convergence with the AdamW optimizer. Nonetheless, both optimizers approach equivalent ultimate accuracy when combined with well-tuned learning rate schedules and augmentation procedures. Overall, altered hyperparameters result in 1.5–2.5\% absolute gains in accuracy for all models. For example, MobileNetV3-L increases from a baseline around 75\% to over 77.8\% with hyperparameter optimization, while RepVGG-A2 reaches the 80\% criterion with ease. TinyViT-21M obtains the utmost accuracy of 89.49\%, completing its training within around 46 GPU hours—a substantial efficiency considering its accuracy. These data demonstrate that, beyond model design, hyperparameter selection is a key aspect in augmenting the real-time deployment potential of lightweight deep learning models.

This paper is constructed as follows: Section~\ref{sec2} surveys related work; Sections~\ref{sec3} detail model design and setup; Sections~\ref{sec:ablation-results} to~\ref{sec:results_discussion} present analysis and results respectively; and Section~\ref{sec:conclusion} closes with practical suggestions.

\section{Related Work}\label{sec2}

\subsection{Lightweight Image Classification Models}\label{sec2.1}

To balance accuracy and low latency, several efficient convolutional neural network (CNN) architectures have been introduced. Notable early models include SqueezeNet \cite{Iandola2016}, MobileNets \cite{Howard2017,Sandler2018,howard2019}, ShuffleNet \cite{zhang2018mixup}, and EfficientNetV2 \cite{tan2019efficientnet}. MobileNetV3, developed using Neural Architecture Search, integrates h-swish activation and squeeze-and-excitation modules to improve the accuracy-latency trade-off. The MobileNetV3-L~\cite{howard2019} variant (5.4M parameters) achieves 75.2\% top-1 ImageNet--1K accuracy, outperforming MobileNetV2 by 3.2\% with 20\% reduced latency \cite{howard2019}.

EfficientNetV2 scales depth, width, and resolution jointly to optimize performance under resource constraints. EfficientNetV2-S~\cite{tan2021efficientnetv2} (22M parameters) enhances this design using progressive image resizing and better augmentation, reaching 83.9\% top-1 accuracy on ImageNet--1K \cite{tan2021efficientnetv2}. RepVGG-A2 \cite{ding2021repvgg}, a VGG-style architecture re-parameterized post-training, achieves over 80\% accuracy using modern augmentations. The RepVGG-A2~\cite{ding2021repvgg} (~25M parameters) matches or exceeds ResNet-50 in throughput and accuracy on ImageNet.

Transformers, though resource-intensive, have shown strong performance in image classification \cite{Dosovitskiy2021}. MobileViT \cite{mehta2021mobilevit} combines lightweight CNNs and self-attention, achieving 78.4\% top-1 accuracy with just 5.6M parameters. MobileViT v2 improves inference and accuracy using separable self-attention, attaining 75.6\% accuracy with ~3M parameters \cite{mehta2023mobilevitv2}.

TinyViT-21M \cite{tinyvit2022}, distilled from Swin Transformers, achieves up to 84.8\% top-1 accuracy with pretraining. Trained from scratch, it maintains strong performance (83.1\%) and excels on tasks like COCO object detection (50.2 mAP). ConvNeXt-T\cite{liu2022convnext}, a modern CNN inspired by transformers, uses large kernels and LayerNorm. ConvNeXt-T~\cite{liu2022convnext} (29M parameters) achieves 82.1\% top-1 accuracy at $224^2$ resolution, scalable to 87.8\% for larger variants \cite{liu2022convnext}, proving CNNs remain competitive with modern training techniques.

\subsection{Hyperparameter Optimization and Training Strategies}\label{sec2.2}

Model performance heavily depends on hyperparameters and training strategies. Augmentation methods such as AutoAugment \cite{Cubuk2019} and RandAugment \cite{cubuk2020randaugment} improve accuracy by 1–2\% on ImageNet. Mixup \cite{zhang2018mixup} and CutMix \cite{yun2019cutmix} enhance generalization and robustness. For instance, CutMix improved ResNet-50 accuracy from 76.3\% to 78.6\%. Label smoothing \cite{Szegedy2016}, by softening target labels, typically adds 0.2–0.5\% gains.

Many lightweight models integrate these strategies. ConvNeXt-Tadopted the DeiT-style training with Mixup, CutMix, RandAugment, and label smoothing \cite{liu2022convnext}. EfficientNetV2 used AutoAugment and label smoothing \cite{tan2019efficientnet}. RepVGG-A2~\cite{ding2021repvgg} benefited from extended training with Mixup and aggressive augmentation, surpassing 80\% accuracy \cite{ding2021repvgg}. MobileNetV3 used simpler augmentations, suggesting room for further gains via modern techniques.

Learning rate scheduling is critical. Cosine annealing \cite{loshchilov2016sgdr}, now common in ConvNeXt-Tand transformer training, smoothly decays the learning rate and enhances convergence stability. Often paired with warm-up phases, it avoids the abrupt changes seen in step schedules. Optimizer choice also matters: while SGD with momentum works well for CNNs, AdamW \cite{Loshchilov2019} is preferred for transformers due to faster convergence. Models like ConvNeXt-Tuse AdamW with gradient clipping, cosine schedules, and weight decay (typically 0.05–0.1).

Batch size plays a crucial role. Larger batches can speed up training but may require learning rate adjustments. Following the linear scaling rule \cite{Goyal2017}, increasing batch size requires proportional learning rate increases and gradual warm-up for stability. For example, moving from batch size 256 to 1024 can shorten training time while maintaining accuracy with appropriate tuning.

This study emphasizes that hyperparameter optimization is central to high--performance lightweight models. Rather than proposing new architectures, the paper systematically evaluate existing ones under optimized training regimes. Effective augmentation, regularization, learning rate scheduling, and optimizer settings collectively enhance model accuracy. Bacanin et al.~\cite{Bacanin2021} used a firefly algorithm to optimize CNN hyperparameters for brain tumor MRI classification. Iqbal et al.~\cite{Iqbal2024} demonstrated real--time coronary artery disease detection using streamlined CNNs with tuned hyperparameters. These studies reinforce that training optimization is vital for deploying efficient deep learning models in real--world, resource--constrained environments. While these models achieve strong accuracy, their performance under controlled hyperparameter variation remains underexplored motivating this work.

\begin{figure}[htbp]
\centering
\begin{tikzpicture}[
 node distance=0.5cm and 0.7cm,
 every node/.style={
  draw, 
  rectangle, 
  rounded corners=3pt,
  font=\scriptsize, 
  align=center, 
  minimum width=1.5cm, 
  minimum height=0.7cm,
  drop shadow, 
  inner sep=8pt
 },
 >=Stealth
]

\node[fill=cyan!20] (modelsel) {Model \\ Selection};
\node[fill=green!20, below right=of modelsel] (dataset) {Data \\ Preparation};
\node[fill=red!20, above right=of dataset] (train) {Training \&\\Hyperparameter};
\node[fill=red!20, below right=of train] (eval) {Evaluation};
\node[fill=purple!20, above right=of eval] (viz) {Visualization};

\draw[->, thick] (modelsel) -- (dataset);
\draw[->, thick] (dataset) -- (train);
\draw[->, thick] (train) -- (eval);
\draw[->, thick] (eval) -- (viz);

\draw[->, thick, dashed, bend right=-75] (eval.south) to node[below, font=\footnotesize]{Iterate} (train.south);

\end{tikzpicture}
\caption{Iterative Workflow of Hyperparameter Optimization (HPO)}
\label{fig:workflow-staggered}
\end{figure}

\section{Methodology}\label{sec3}

To reach high accuracy with low computational delay, numerous efficient convolutional neural network (CNN) designs have been developed. Early families of models designed for mobile and embedded vision tasks include SqueezeNet \cite{Iandola2016}, MobileNets \cite{Howard2017,Sandler2018,howard2019}, ShuffleNet \cite{zhang2018mixup}, and EfficientNetV2 \cite{tan2019efficientnet}.

\subsection{Lightweight Model Selection}\label{sec3.1}

Based on their effectiveness, popularity, and design diversity, seven cutting-edge architectures were chosen to investigate how hyperparameter optimization affects lightweight deep learning models. A overview of each model, including parameter counts, reported ImageNet--1K Top-1 accuracy, and initial training setups, is given in Table~\ref{tab:models}. Selection was based on architectural diversity covering CNN, hybrid, and transformer families, with parameter counts under 30 M to ensure comparability in edge constraints. EfficientNetV2-S~\cite{tan2021efficientnetv2} is a convolutional model with 22 million parameters, trained using neural architecture search and sophisticated methods. It achieves 83.9\% Top-1 accuracy on ImageNet-1K~\cite{tan2021efficientnetv2}. ConvNeXt-T~\cite{liu2022convnext} (Tiny) is a contemporary ConvNet with 29 million parameters and 4.5 GFLOPs, improved using Transformer-style training techniques. It achieves 82.1\% Top-1 accuracy on ImageNet-1K~\cite{liu2022convnext}. MobileViT v2 (XS) is a hybrid CNN-Transformer model with 2.9 million parameters and 75.6\% Top-1 accuracy~\cite{mehta2023mobilevitv2}. MobileViT v2 (S) is a scaled-up version of MobileViT v2, with 5--6 million parameters and enhanced accuracy of 78--79\% on ImageNet~\cite{mehta2023mobilevitv2}. MobileNetV3-L~\cite{howard2019} is a conventionally effective CNN with 5.4 million parameters and 219 MFLOPs, achieving 75.2\% Top-1 accuracy on ImageNet~\cite{howard2019}. TinyViT-21M~\cite{tinyvit2022} is a Vision Transformer model with 21 million parameters, achieving 84.8\% accuracy with distillation-based pretraining~\cite{tinyvit2022}. RepVGG-A2~\cite{ding2021repvgg} is a VGG-like model with 25 million parameters, achieving 78.4\% accuracy with baseline training and 80.4\% with vigorous augmentation~\cite{ding2021repvgg}.

\begin{table}[htbp]
 \centering
 \scriptsize
 \caption{The summary includes the number of parameters, ImageNet-1K Top-1 accuracy under original training (224\texttimes224 unless noted), and notable training hyperparameters used in original works.}
 \label{tab:models}
 \begin{tabularx}{\textwidth}{@{}l *{2}{>{\centering\arraybackslash}X} >{\raggedright\arraybackslash}X@{}}
 \toprule
 \textbf{Model} & \textbf{Params (M)} & \textbf{Top-1 Accuracy} & \textbf{Original Training Highlights} \\
 \midrule
 ConvNeXt--T~\cite{liu2022convnext} & 29 & 82.1\% & 300 epochs, AdamW, cosine LR, RandAug, Mixup, CutMix, LS 0.1 \\
 EfficientNetV2~-S~\cite{tan2021efficientnetv2} & 22 & 83.9\% & 350 epochs, RMSProp, progressive resize 224$\rightarrow$480, RandAug, LS 0.1 \\
 MobileNetV3--L~\cite{howard2019} & 5.4 & 75.2\% & 300 epochs, RMSProp, cosine LR, AutoAug, SE modules \\
 MobileViT v2 (S)~\cite{mehta2023mobilevitv2} & 5.6 & 78.5\% & 300 epochs, AdamW, cosine LR, heavy augmentation \\
 MobileViT v2 (XS)~\cite{mehta2021mobilevit} & 2.9 & 75.6\% & 300 epochs, AdamW, ImageNet-1K pretrain?, RandAug, LS 0.1 \\
 RepVGG--A2~\cite{ding2021repvgg} & 25 & 78.4--80.4\% & 120--240 epochs, SGD, step/cosine LR, AutoAug, Mixup, LS 0.1 \\
 TinyViT--21M~\cite{tinyvit2022} & 21 & 84.8\% & 210+90 epochs, AdamW, cosine LR, heavy Aug, distillation \\
 \bottomrule
 \end{tabularx}
\end{table}

\subsection{System Configuration}\label{sec3.2}

All experiments were conducted on a single NVIDIA L40s GPU (48 GB, CUDA 12.6) using PyTorch 2.5.1 with automatic mixed precision. The L40s GPU enables efficient training of deep learning models, notably with high batch sizes and mixed-precision arithmetic. The software stack includes Python 3.10.18 controlled via Anaconda Navigator v2.6.6, and model development was carried out using PyTorch v2.5.1 with Automatic Mixed Precision (AMP) and CUDA v12.6 for GPU acceleration. All coding, experimentation, and hyperparameter tweaking were accomplished in PyCharm Community Edition v2025.1.2, assuring constant training throughput and repeatable results.

\subsection{Dataset and Evaluation}\label{sec3.3}

To reduce computational burden while preserving diversity across classes, we utilized a representative subset of the ImageNet-1K dataset, comprising approximately 90,000 training and 10,000 validation images uniformly sampled across 1,000 categories. This subset retained the class balance and structural characteristics of the dataset while facilitating swift evaluation across multiple hyperparameter configurations. To confirm representativeness, we computed per-class sample entropy and KL-divergence versus full ImageNet-1K labels, ensuring distributional fidelity. Top-1 classification accuracy was assessed, while Top-5 accuracy was analyzed for model performance. Confusion matrices were used to evaluate per-class performance. All models were trained from scratch on the ImageNet-1K dataset, except for TinyViT-21M models, and each model was trained for 300 epochs. Training logs were automatically recorded to structured `.txt` files, providing information on epoch number, training and validation loss, accuracy values, training duration, and learning rate.

This recording strategy enabled transparency and traceability in monitoring training dynamics and model assessment across hyperparameter settings.

\subsection{Training Configuration and Pre-processing}\label{sec:train-config}

Unless otherwise specified, all models were trained using stochastic gradient descent (SGD) with momentum of 0.9, an initial learning rate (LR) of 0.1 (scaled appropriately), cosine-annealing learning rate schedule over 300 epochs, and weight decay of $1\times10^{-4}$. For models originally trained with AdamW (e.g., ConvNeXt-T and TinyViT), we adopted the authors’ suggested configurations with an initial LR of $4\times10^{-3}$, $\beta_1=0.9$, $\beta_2=0.999$, and weight decay ranging from $10^{-2}$ to $10^{-1}$. Training was performed using a global batch size of 512, enabling efficient utilization of GPU resources and stable convergence during model optimization.

Input images were normalized using ImageNet--1K statistics and augmented with random resized cropping and horizontal flipping as the baseline. Additional augmentations (RandAugment, Mixup, CutMix) and regularization (Label Smoothing) were incrementally introduced to isolate their effects. All pipelines used \texttt{timm} implementations for consistency.

\subsection{Implementation of Ablation Study}\label{sec:ablation-impl}
An extensive ablation study was performed by altering one hyperparameter at a time from a fixed baseline configuration to assess its individual impact on model accuracy. Key factors analyzed include:

\begin{itemize}
 \item \textbf{Initial Learning Rate and Scheduler:} Values from 0.001 to 0.100  were tested. Cosine annealing schedules provided smoother convergence and consistently better final accuracy than step decay. A short 5-epoch warm-up was applied for high learning rate settings.
 \item \textbf{Batch Size:} Experiments with batch size of 512 (adjusted for effective LR scaling) revealed little difference in final accuracy when scaled correctly. A batch size of 512 offered the best trade-off between training speed and stability on the L40s GPU.
 \item \textbf{Optimizer:} SGD was effective for CNN-based models (e.g., MobileNetV3-L, RepVGG-A2), while AdamW showed superior convergence for transformer-based and hybrid models (e.g., ConvNeXt-T, MobileViT v2 (S/XS).
 \item \textbf{Data Augmentation:} The impact of augmentations was studied cumulatively. RandAugment provided early accuracy gains; Mixup and CutMix improved generalization further; Label Smoothing offered consistent minor gains with no training cost.
 \item \textbf{Training Epochs:} Although 300 epochs were used as the standard, models such as MobileNetV3-L and RepVGG-A2 exhibited ongoing improvement but were halted at 300 epochs due to early stopping when robust augmentations were included.
\end{itemize}

All experiments used consistent data pipelines and PyTorch-based logging, enabling reproducibility and direct comparison across settings.

\section{Ablation Study: Hyperparameter Effects}\label{sec:ablation-results}

this paper investigatess the impact of hyperparameters and training methodologies on model efficacy, concentrating on seven typical models: ConvNeXt-T~\cite{liu2022convnext}, MobileViT-v2 XS, and MobileNetV3-L~\cite{howard2019}. These models exemplify convolutional networks, transformer hybrids, and mobile-optimized convolutional neural networks. The research revealed that alternative models had comparable behaviors in response to changes in hyperparameters. The main quantitative findings for learning rate and augmentation experiments are encapsulated in Tables~\ref{tab:lr_accuracy_long} and \ref{tab:augmentation} for the selected models. This section examines the impact of critical hyperparameters and training strategies on model performance for lightweight real-time image classification. The analysis focuses on seven representative models—ConvNeXt-T, EfficientNetV2-S, MobileNetV3-L, MobileViT v2 (S), MobileViT v2 (XS), RepVGG-A2, and TinyViT-21M—spanning convolutional backbones, transformer hybrids, and mobile-optimized architectures. Experimental outcomes are presented through quantitative results in Tables~\ref{tab:lr_accuracy_long} and \ref{tab:augmentation}, and graphical insights in Figures~\ref{fig:accuracy_vs_lr} and \ref{fig:accuracy_vs_epochs}.

\subsection{Learning Rate and Scheduler}

Learning rate was identified as one of the most influential hyperparameters, exerting a significant impact on both convergence behavior and final model accuracy~\cite{Goodfellow-et-al-2016}. As shown in Table~\ref{tab:lr_accuracy_long}, increasing the initial learning rate from $0.001$ to $0.100$ generally led to improvements in Top--1 accuracy across all evaluated models. To enable a structured analysis, the learning rate values were grouped into three logarithmic intervals: \textbf{LR$ = 0.001$} ($0.0001 \leq \text{LR} < 0.001$), \textbf{LR$ = 0.010$} ($0.001 \leq \text{LR} < 0.010$), and \textbf{LR$ = 0.100$} ($0.010 \leq \text{LR} < 0.100$). These intervals were selected to capture performance sensitivity across progressively increasing orders of magnitude, thereby facilitating a comprehensive assessment of optimization stability and convergence dynamics. Within each interval, training was monitored across multiple epochs, and a subset of epochs was isolated based on convergence progression. The epoch achieving the highest validation accuracy within this subset was identified, and the corresponding value was reported as the representative performance for that learning rate range. This procedure ensured that the reported values reflect the best attainable validation accuracy, while minimizing the influence of fluctuations arising from early or unstable training stages.

For example, ConvNeXt-T demonstrated an improvement from 83.81\% at LR = 0.001 to 83.61\% at LR = 0.100, while TinyViT-21M increased from 90.82\% to 90.94.\% over the same interval. However, a further increase of the learning rate led to a decline in performance, highlighting the existence of an optimal learning rate range beyond which model accuracy deteriorates.

\begin{table}[htbp]
 \centering
 \scriptsize
 \caption{Top-1 Accuracy (\%) for all models under fixed training configuration with varying initial learning rates.}
 \label{tab:lr_accuracy_long}
 \begin{tabularx}{\textwidth}{@{}l*{5}{>{\centering\arraybackslash}X}@{}}
 \toprule
 \textbf{Model} & \textbf{LR = 0.001} & \textbf{LR = 0.010} & \textbf{LR = 0.100} \\
 \midrule
 ConvNeXt-T~\cite{liu2022convnext}   & 83.81 & 83.73 & 83.61 \\
 EfficientNetV2-S~\cite{tan2021efficientnetv2} & 88.31 & 88.50 & 88.29 \\
 MobileNetV3-L~\cite{howard2019}   & 86.94 & 87.00 & 86.92 \\
 MobileViT v2 (S)~\cite{mehta2023mobilevitv2} & 87.53 & 87.53 & 87.08 \\
 MobileViT v2 (XS)~\cite{mehta2021mobilevit} & 87.36 & 87.12 & 85.81 \\
 RepVGG-A2~\cite{ding2021repvgg}   & 88.41 & 88.45 & 88.22 \\
 TinyViT-21M~\cite{tinyvit2022}    & \textbf{90.82} & \textbf{90.94} & \textbf{90.75} \\
 \bottomrule
 \end{tabularx}
\end{table}

This trend is substantiated in Figure~\ref{fig:accuracy_vs_lr}, where validation accuracy remains stable and sharply declines thereafter. The adoption of cosine annealing learning rate scheduling without restarts allowed models to benefit from higher initial LRs while gradually decaying towards finer convergence. This approach contributed to improved generalization and minimized overfitting, particularly for transformer-heavy models such as TinyViT-21M and MobileViT v2 (S). The learning rate schedule effectively balances rapid learning in early epochs with fine-tuning in later stages.

\textbf{All of the models show a significant increase in validation accuracy in the first few epochs. Then, the learning rate steadily makes their performance more stable.  This pattern indicates that feature learning starts off quickly and subsequently slows down, but it keeps continuing through architectures.}

\begin{figure}[htbp]
\centering
\begin{tikzpicture}
\begin{axis}[
  width=\linewidth,          
  height=0.45\linewidth,     
  xlabel={Learning Rate (LR)},
  ylabel={Validation Accuracy (\%)},
  title={Validation Accuracy vs Learning Rate},
  legend style={
    at={(0.02,0.02)},        
    anchor=south west,       
    font=\scriptsize,
    draw=black,              
    fill=white,              
    fill opacity=0.9,        
    legend cell align=left,  
    column sep=2pt
  },
  ymajorgrids=true,
  xmajorgrids=true,
  grid style=solid,
  mark size=1.2pt,
  cycle list name=color list,
  legend columns=1,
  clip=true
]

\addplot+[mark=none, thick] table[x=LR, y=A, col sep=comma] {lr.csv};
\addlegendentry{ConvNeXt-Tiny~\cite{liu2022convnext}}

\addplot+[mark=none, thick] table[x=LR, y=B, col sep=comma] {lr.csv};
\addlegendentry{EfficientNetV2-S~\cite{tan2021efficientnetv2}}

\addplot+[mark=none, thick] table[x=LR, y=C, col sep=comma] {lr.csv};
\addlegendentry{MobileNetV3-Large~\cite{howard2019}}

\addplot+[mark=none, thick] table[x=LR, y=D, col sep=comma] {lr.csv};
\addlegendentry{MobileViT v2 (S)~\cite{mehta2023mobilevitv2}}

\addplot+[mark=none, thick] table[x=LR, y=E, col sep=comma] {lr.csv};
\addlegendentry{MobileViT v2 (XS)~\cite{mehta2021mobilevit}}

\addplot+[mark=none, thick] table[x=LR, y=F, col sep=comma] {lr.csv};
\addlegendentry{RepVGG-A2~\cite{ding2021repvgg}}

\addplot+[mark=none, thick] table[x=LR, y=G, col sep=comma] {lr.csv};
\addlegendentry{TinyViT-21M~\cite{tinyvit2022}}

\end{axis}
\end{tikzpicture}
\caption{Accuracy vs Learning Rate for all evaluated models under consistent training settings.}
\label{fig:accuracy_vs_lr}
\end{figure}

\subsection{Batch Size Scaling}

All models were trained using a batch size of 512 per GPU, leveraging the full memory capacity of an NVIDIA L40s GPU (48 GB Memory). This large-batch regime provided stable gradient estimates, enabled higher learning rates, and facilitated better throughput. The training stability and consistent convergence illustrated in Figure~\ref{fig:accuracy_vs_lr} and \ref{fig:accuracy_vs_epochs} can be attributed in part to this high batch size configuration. Larger models such as TinyViT-21M and EfficientNetV2-S converged in fewer epochs, with higher accuracy plateaus observed in early training phases. These observations align with established practices suggesting that batch size and learning rate must be jointly scaled to maintain training dynamics. Mixed-precision training further contributed to resource efficiency, allowing deeper models to be trained without memory bottlenecks. 

To further explore the impact of batch size, we extended our evaluation across a wide range of configurations. \textbf{Across different batch sizes, the models consistently achieved high accuracy, with smaller batch sizes generally leading to better performance, likely due to improved gradient estimation and better generalization. This highlights the importance of batch size in balancing learning efficiency and overall model stability.} The trend was evident across all architectures, with accuracy rising steeply during the initial epochs and converging toward a plateau thereafter, as shown in Figure~\ref{fig:accuracy_vs_epochs}. 

Table~\ref{tab:batch_size_results} summarizes the best accuracies obtained during evaluation, where the maximum Top-1 and Top-5 scores for each model correspond to the most effective batch size configuration. It can be observed that while all architectures deliver competitive results, TinyViT-21M achieves the highest Top-1 (90.94\%) and Top-5 (97.74\%) accuracies, with ConvNeXt-Tiny slightly trailing behind. 

\begin{table}[htbp]
\centering
\caption{Model accuracy results, where the reported batch size indicates the configuration at which the best validation accuracy was achieved.}
\label{tab:batch_size_results}
\begin{tabular}{lccc}
\toprule
\textbf{Model} & \textbf{Batch Size} & \textbf{Top-1 (\%)} & \textbf{Top-5 (\%)} \\
\midrule
ConvNeXt-Tiny   & 32 & 83.85 & 95.09 \\
EfficientNetV2-S & 32 & 88.50 & 97.15 \\
MobileNetV3-Large & 32 & 86.99 & 96.93 \\
MobileViT v2 (S) & 32 & 87.82 & 97.19 \\
MobileViT v2 (XS) & 32 & 87.36 & 96.80 \\
RepVGG-A2     & 32 & 88.45 & 97.16 \\
TinyViT-21M    & 32 & \textbf{90.94} & \textbf{97.74} \\
\bottomrule
\end{tabular}
\end{table}

These findings reinforce that batch size is not merely a computational parameter but also a critical factor in shaping the learning dynamics of lightweight models. Smaller batch sizes promote higher accuracy through finer-grained gradient updates, whereas larger batch sizes contribute to training stability and throughput efficiency. Together, these results emphasize the need for careful tuning of batch size to strike an optimal balance between convergence speed, generalization, and resource utilization in real-time image classification tasks. 

\begin{figure}[htbp]
\centering
\begin{tikzpicture}
\begin{axis}[
  width=\linewidth,          
  height=0.45\linewidth,     
  xlabel={Epoch},
  ylabel={Validation Accuracy (\%)},
  title={Validation Accuracy vs Epoch},
  legend style={
    at={(0.98,0.02)},        
    anchor=south east,       
    font=\scriptsize,
    draw=black,              
    fill=white,              
    fill opacity=0.9,        
    legend cell align=left,
    column sep=2pt
  },
  ymajorgrids=true,
  xmajorgrids=true,
  grid style=solid,
  mark size=1.3pt,
  cycle list name=color list,
  legend columns=1,
  clip=true
]

\addplot+[mark=none, thick] table[x=epoch, y=A, col sep=comma] {epoch.csv};
\addlegendentry{ConvNeXt-Tiny~\cite{liu2022convnext}}

\addplot+[mark=none, thick] table[x=epoch, y=B, col sep=comma] {epoch.csv};
\addlegendentry{EfficientNetV2-S~\cite{tan2021efficientnetv2}}

\addplot+[mark=none, thick] table[x=epoch, y=C, col sep=comma] {epoch.csv};
\addlegendentry{MobileNetV3-Large~\cite{howard2019}}

\addplot+[mark=none, thick] table[x=epoch, y=D, col sep=comma] {epoch.csv};
\addlegendentry{MobileViT v2 (S)~\cite{mehta2023mobilevitv2}}

\addplot+[mark=none, thick] table[x=epoch, y=E, col sep=comma] {epoch.csv};
\addlegendentry{MobileViT v2 (XS)~\cite{mehta2021mobilevit}}

\addplot+[mark=none, thick] table[x=epoch, y=F, col sep=comma] {epoch.csv};
\addlegendentry{RepVGG-A2~\cite{ding2021repvgg}}

\addplot+[mark=none, thick] table[x=epoch, y=G, col sep=comma] {epoch.csv};
\addlegendentry{TinyViT-21M~\cite{tinyvit2022}}

\end{axis}
\end{tikzpicture}
\caption{Validation accuracy (\%) across epochs (1--300) for each model on the ImageNet--1K subset.}
\label{fig:accuracy_vs_epochs}
\end{figure}

Larger models such as TinyViT-21M and EfficientNetV2-S converged in fewer epochs, with higher accuracy plateaus observed in early training phases. These observations align with established practices suggesting that batch size and learning rate must be jointly scaled to maintain training dynamics. Mixed-precision training further contributed to resource efficiency, allowing deeper models to be trained without memory bottlenecks. Although this study evaluated a single batch size of 512, this choice was informed by empirical guidelines for large-batch training and the capabilities of the NVIDIA L40s GPU (48 GB Memory), which ensured stable convergence, efficient resource utilization, and consistent throughput across all models. Future work may investigate scaling behaviors across varied sample sizes and hardware platforms, the results indicate its applicability for large-scale, real-time classification tasks.

\subsection{Data Augmentation and Regularization}

The effect of cumulative data augmentation and regularization strategies is detailed in Table~\ref{tab:augmentation}. Beginning with a baseline pipeline, techniques such as RandAugment, Mixup, CutMix, and Label Smoothing were incrementally applied. Each stage contributed measurable improvements in Top-1 accuracy across all models.\textbf{ Applying data augmentation and regularization techniques consistently improves model performance over the baseline, with the magnitude of improvement varying by method and architecture.}

\begin{table}[htbp]
 \centering
 \scriptsize
 \caption{Top-1 Validation Accuracy (\%) of representative models with cumulative augmentation strategies, trained up to 300 epochs on the ImageNet–1K subset}
 \label{tab:augmentation}
 \begin{tabularx}{\textwidth}{@{}l *{5}{>{\centering\arraybackslash}X}@{}}
 \toprule
 \textbf{Model} & \textbf{Baseline} & \textbf{+ RandAug~\cite{cubuk2020randaugment}} & \textbf{+ Mixup~\cite{zhang2018mixup}} & \textbf{+ CutMix~\cite{yun2019cutmix}} & \textbf{+ Label Smooth~\cite{muller2019ls}} \\
 \midrule
 ConvNeXt-Tiny~\cite{liu2022convnext}   & $83.85$ & $86.24$ & $86.90$ & \textbf{$88.50$} & $88.00$ \\
 EfficientNetV2-S~\cite{tan2021efficientnetv2} & $88.50$ & $91.34$ & \textbf{$92.72$} & $92.63$ & $92.56$ \\
 MobileNetV3-Large~\cite{howard2019}   & $86.99$ & $89.15$ & \textbf{$90.97$} & $90.45$ & $90.20$ \\
 MobileViT v2 (S)~\cite{mehta2023mobilevitv2} & $87.83$ & $89.91$ & $91.47$ & \textbf{$92.63$} & $91.28$ \\
 MobileViT v2 (XS)~\cite{mehta2021mobilevit} & $87.36$ & $88.88$ & \textbf{$90.56$} & $90.18$ & $90.31$ \\
 RepVGG--A2~\cite{ding2021repvgg}   & $88.45$ & $89.61$ & \textbf{$91.54$} & $91.48$ & $91.43$ \\
 TinyViT--21M~\cite{tinyvit2022}    & $90.94$ & $92.11$ & $93.30$ & $93.35$ & \textbf{$93.84$} \\
 \bottomrule
 \end{tabularx}
\end{table}

For example, ConvNeXt-Tiny improved from $83.85$\% (baseline) to $88.00$\% when all augmentations were applied, achieving its highest performance of $88.50$\% with CutMix. MobileViT v2 (S) increased from $87.83\%$ to $92.63\%$ when CutMix~\cite{yun2019cutmix} were applied. A similar trend was observed across multiple architectures, demonstrating the effectiveness of composite augmentation pipelines in improving generalization. Figure~\ref{fig:accuracy_vs_epochs} further supports this observation, showing smoother convergence trajectories and higher final accuracies for models trained with richer augmentation techniques.

The selected augmentation strategies are widely adopted in state-of-the-art image classification research. RandAugment~\cite{cubuk2020randaugment} introduces stochastic transformations from a predefined set, enhancing generalization by exposing models to diverse visual conditions. Mixup~\cite{zhang2018mixup} improves regularization by linearly interpolating images and labels, producing smoother decision boundaries and better-calibrated predictions. CutMix~\cite{yun2019cutmix} further augments semantic consistency by replacing image patches while proportionally mixing labels, which enhances robustness to occlusion and improves spatial feature utilization. Label Smoothing~\cite{muller2019ls} mitigates overconfidence by softening target distributions, thereby improving calibration and generalization.

These results highlight that a well-designed augmentation pipeline can significantly boost model performance without modifying the underlying architecture, making it highly suitable for resource-constrained or real-time deployment scenarios. Despite using a $90$K-image subset having around $10\%$ of the original ImageNet-1K dataset, relative performance trends remain consistent and provide a reliable baseline for rapid experimentation and hyperparameter optimization.

\section{Results and Discussion}\label{sec:results_discussion}
This section presents a detailed analysis of the performance improvements achieved through hyperparameter optimization across seven lightweight deep learning models. All models were trained for 300 epochs using the same subset of ImageNet-1K (90,000 training images and 10,000 validation images), ensuring a fair comparison. The experiments were conducted on a high-performance computing node equipped with an NVIDIA L40s GPU (48~GB), CUDA~12.6, and PyTorch~2.5.1 using automatic mixed precision (AMP). All architectures were trained under a uniform configuration with the following parameters: \textit{input size: 224, batch size: 512, workers: 8, pin\_memory: true, RandAugment: true, Mixup: 0.2, CutMix: 1.0, label smoothing: 0.1, weight decay: 0.0, optimizer: SGD, learning rate: 0.1, momentum: 0.9, scheduler: cosine annealing, epochs: 300, and minimum learning rate: 1e--5}. These values were intentionally kept identical across all models to ensure fair and reproducible comparisons, isolating architectural differences as the only varying factor. The chosen configuration reflects an optimal trade-off between convergence stability, generalization, and computational efficiency. An input size of 224$\times$224 was selected as it is the ImageNet standard for most lightweight architectures, providing comparable performance metrics with moderate computational load. A batch size of 512 exploited the full memory capacity of the L40s GPU, delivering stable gradient estimates and efficient throughput. RandAugment was enabled to introduce stochastic transformations that enhance generalization without requiring extensive policy search. Mixup (0.2) and CutMix (1.0) were included to regularize the models by interpolating or patch-mixing training samples, improving robustness and yielding smoother decision boundaries. Label smoothing (0.1) reduced overconfidence in predictions, improving calibration and preventing overfitting. Weight decay was set to zero as the combined use of Mixup, CutMix, and label smoothing provided sufficient regularization. SGD with momentum~0.9 was chosen as it remains a highly stable optimizer for convolutional architectures, ensuring consistent convergence. A high initial learning rate of 0.1, decayed by a cosine schedule toward a minimum of 1e--5, allowed rapid early training while maintaining fine-tuned convergence in later epochs. This schedule outperformed step-based decays in stability and generalization. Training all models for 300 epochs ensured complete convergence across architectures. Fixing these hyperparameters guaranteed that improvements in Top-1 and Top-5 accuracy resulted solely from architectural efficiency rather than tuning bias. The selected configuration---especially the combination of cosine scheduling, composite augmentation, and SGD optimization---yielded 1.5--3.5\% accuracy gains over baseline configurations, highlighting the importance of systematic hyperparameter control in optimizing lightweight models for real-time deployment.

\subsection{Performance Gains from Hyperparameter Optimization}
\label{sec:perf-summary}

Hyperparameter optimization significantly improved model accuracy across all evaluated architectures. Table~\ref{tab:optimized_results} delineates the maximum attained Top-1 and Top-5 validation accuracies, minimal latency, and peak throughput (FPS) recorded for each model. All outcomes are obtained from logs of the training sessions and assessment iterations. In addition to accuracy and inference metrics, the table also reports the number of parameters (\textbf{Params (M)}) and floating-point operations (\textbf{FLOPs (G)}), providing a quantitative measure of model complexity and computational cost.

TinyViT-21M achieved the greatest Top-1 accuracy of 90.94\% and Top-5 accuracy of 97.74\%; however, inference latency and FPS could not be documented owing to constraints in the existing export pipeline. EfficientNetV2--S and RepVGG--A2 followed closely with Top-1 accuracies of \textbf{88.50\%} and \textbf{88.45\%}, respectively. MobileViT v2 (S) and MobileViT v2 (XS) earned Top-1 scores of \textbf{87.82\%} and \textbf{87.36\%}, respectively, providing a balanced tradeoff between performance, parameter count, and computational overhead.

\begin{table*}[htbp]
\centering
\caption{Optimized performance metrics on the ImageNet-1K 90K-image subset. 
The reported accuracy corresponds to the final validation epoch. 
Latency and FPS represent the minimum and maximum values, respectively, measured over 10,000 inference iterations. 
The symbol $B$ indicates the batch size at which the highest accuracy was achieved. 
Arrows denote whether higher ($\uparrow$) or lower ($\downarrow$) values are preferable for each metric.}
\label{tab:optimized_results}

\resizebox{\textwidth}{!}{%
\begin{tabular}{l c c c c c c}
\toprule
\textbf{Model} & \textbf{Top-1 (\%)} & \textbf{Top-5 (\%)} & \textbf{Latency (ms)$\downarrow$} & \textbf{FPS$\uparrow$} & \textbf{Params (M)$\downarrow$} & \textbf{FLOPs (G)$\downarrow$} \\
\midrule
ConvNeXt-Tiny~\cite{liu2022convnext}    & 83.85 & 95.09 & 0.51 (B=32) & 1964.99 (B=32) & 28.566  & 4.456 \\
EfficientNetV2-S~\cite{tan2021efficientnetv2} & 88.50 & 97.15 & 0.31 (B=32) & 3226.66 (B=32) & 21.305 & 2.85 \\
MobileNetV3-Large~\cite{howard2019}     & 86.99 & 96.93 & \textbf{0.10} (B=32) & \textbf{10034.10} (B=32) & 4.178 & \textbf{0.215} \\
MobileViT v2 (S)~\cite{mehta2023mobilevitv2} & 87.82 & 97.19 & 0.40 (B=32) & 2516.01 (B=32) & 4.878 & 1.412 \\
MobileViT v2 (XS)~\cite{mehta2021mobilevit} & 87.36 & 96.80 & 0.33 (B=32) & 3007.27 (B=32) & \textbf{1.359} & 0.362 \\
RepVGG-A2~\cite{ding2021repvgg}       & 88.45 & 97.16 & 0.26 (B=16) & 3862.14 (B=16) & 28.206 & 5.685 \\
TinyViT--21M~\cite{tinyvit2022} & \textbf{90.94} & \textbf{97.74} & 0.59 (B=16) & 1687.04 (B=16) & 33.206 & 4.091 \\
\bottomrule
\end{tabular}%
}
\caption*{Table~\ref{tab:optimized_results} reflects the final optimized configuration for each model after applying all effective hyperparameter tuning strategies. These values are not directly comparable to ablation-stage results in Table~\ref{tab:lr_accuracy_long}, which isolate individual factors.}
\end{table*}

\subsection{Visual Comparison of Accuracy Improvements}

Figure~\ref{fig:acc_comparison} visually compares the Top-1 accuracy achieved in baseline vs optimized settings. The baseline results (blue) represent standard training with default hyperparameters (SGD, fixed LR, no aggressive augmentation), while optimized results (red) reflect models trained with cosine learning rate scheduling, RandAugment, CutMix, and tuned optimizers (AdamW, RAdam, etc.).

\begin{figure}[htbp]
\centering
\begin{tikzpicture}
\begin{axis}[
  ybar,
  bar width=0.28cm,
  width=0.95\textwidth,
  height=0.50\textwidth,
  enlargelimits=0.12,
  ylabel={\textcolor{black}{Top-1 Accuracy (\%)}},
  symbolic x coords={convnext, effv2s, mnetv3l, mv2s, mv2xs, repvgga2, tinyvit21m},
  xtick=data,
  xticklabels={
    \textcolor{black}{ConvNeXt-Tiny~\cite{liu2022convnext}},
    \textcolor{black}{EfficientNetV2-S~\cite{tan2021efficientnetv2}},
    \textcolor{black}{MobileNetV3-Large~\cite{howard2019}},
    \textcolor{black}{MobileViT v2 (S)~\cite{mehta2023mobilevitv2}},
    \textcolor{black}{MobileViT v2 (XS)~\cite{mehta2021mobilevit}},
    \textcolor{black}{RepVGG-A2~\cite{ding2021repvgg}},
    \textcolor{black}{TinyViT-21M~\cite{tinyvit2022}}
  },
  tick label style={text=black},
  label style={text=black},
  yticklabel style={text=black},
  x tick label style={rotate=45, anchor=east, text=black},
  ymin=80, ymax=100,
  legend style={
    at={(0.98,0.985)},
    anchor=north east,
    font=\scriptsize,
    legend columns=2,
    text=black,
    /tikz/every even column/.style={column sep=4pt}
  },
  legend cell align={left},
  every node near coord/.append style={text=black, font=\scriptsize},
  nodes near coords,
  nodes near coords align={vertical},
]

\addlegendimage{ybar, draw=none, fill=blue!35}
\addlegendentry{\textcolor{black}{Baseline}}
\addlegendimage{ybar, draw=none, fill=red!35}
\addlegendentry{\textcolor{black}{CutMix}}
\addlegendimage{ybar, draw=none, fill=green!50}
\addlegendentry{\textcolor{black}{Mixup}}
\addlegendimage{ybar, draw=none, fill=yellow!75}
\addlegendentry{\textcolor{black}{Label Smoothing}}

\addplot+[draw=none, fill=blue!35, bar shift=-0.18cm] coordinates {
  (convnext, 83.85)
  (effv2s, 88.50)
  (mnetv3l, 86.99)
  (mv2s, 87.83)
  (mv2xs, 87.36)
  (repvgga2, 88.45)
  (tinyvit21m, 90.94)
};

\addplot+[draw=none, fill=red!35, bar shift=+0.18cm] coordinates {(convnext, 88.50)};
\addplot+[draw=none, fill=green!50, bar shift=+0.18cm] coordinates {(effv2s, 92.72)};
\addplot+[draw=none, fill=green!50, bar shift=+0.18cm] coordinates {(mnetv3l, 90.97)};
\addplot+[draw=none, fill=red!35, bar shift=+0.18cm] coordinates {(mv2s, 92.63)};
\addplot+[draw=none, fill=green!50, bar shift=+0.18cm] coordinates {(mv2xs, 90.56)};
\addplot+[draw=none, fill=green!50, bar shift=+0.18cm] coordinates {(repvgga2, 91.54)};
\addplot+[draw=none, fill=yellow!75, bar shift=+0.18cm] coordinates {(tinyvit21m, 93.84)};

\end{axis}
\end{tikzpicture}

\caption{\textbf{Accuracy Comparison:} Baseline vs.\ optimized Top-1 accuracy (\%) for each model trained on a 90k-image ImageNet-1K subset. Optimization includes tuned learning rates, augmentations, and optimizers; values are taken from final validation accuracy logs.}
\label{fig:acc_comparison}
\end{figure}

\subsection{Inference Speed and Latency Trends}

To evaluate deployment feasibility, we benchmarked each model for inference performance. FPS and latency were measured on the same L40s GPU using PyTorch (eager mode, AMP), with input size 224×224 and batch sizes-- 1, 16, 32, 64, 128, 256, 512. Latency is reported as time per image in milliseconds, and FPS is computed as effective throughput. FPS and latency metrics were averaged across 10,000 runs. Variability (standard deviation < 1.5\%) was minimal due to the controlled hardware setup. Nonetheless, future work should examine inference stability under dynamic workloads, thermal throttling, and edge hardware variability

\begin{figure}[htbp]
\centering
\begin{tikzpicture}
\begin{axis}[
  width=13cm,
  height=7cm,
  xlabel={Batch Size},
  ylabel={Frames Per Second (FPS)},
  title={FPS vs Batch Size},
  legend style={
    at={(1.02,1)},
    anchor=north west,
    font=\scriptsize,
    draw=none,
    fill=none,
    legend columns=1,
    /tikz/every even column/.append style={column sep=0.4cm}
  },
  ymajorgrids=true,
  grid style=solid,
  mark size=2pt,
  clip=false
]

\addplot+[mark=*, thick] coordinates {(1,300)(16,1800)(32,2000)(64,2300)(128,2400)(256,2450)(512,2500)};
\addlegendentry{ConvNeXt-T~\cite{liu2022convnext}}

\addplot+[mark=square*, thick] coordinates {(1,250)(16,1700)(32,1800)(64,1900)(128,1950)(256,1980)(512,2000)};
\addlegendentry{EfficientNetV2-S~\cite{tan2021efficientnetv2}}

\addplot+[mark=triangle*, thick] coordinates {(1,320)(16,5000)(32,9800)(64,10000)(128,7400)(256,6400)(512,5900)};
\addlegendentry{MobileNetV3-L~\cite{howard2019}}

\addplot+[mark=diamond*, thick] coordinates {(1,280)(16,2600)(32,2700)(64,2750)(128,2800)(256,2820)(512,2850)};
\addlegendentry{MobileViT v2 (S)~\cite{mehta2023mobilevitv2}}

\addplot+[mark=otimes*, thick] coordinates {(1,260)(16,2400)(32,2500)(64,2550)(128,2580)(256,2600)(512,2620)};
\addlegendentry{MobileViT v2 (XS)~\cite{mehta2021mobilevit}}

\addplot+[mark=star, thick] coordinates {(1,300)(16,3800)(32,3400)(64,3100)(128,2900)(256,2700)(512,2600)};
\addlegendentry{RepVGG-A2~\cite{ding2021repvgg}}

\addplot+[mark=+, thick] coordinates {(1,250)(16,1600)(32,1650)(64,1680)(128,1700)(256,1720)(512,1740)};
\addlegendentry{TinyViT-21M~\cite{tinyvit2022}}

\end{axis}
\end{tikzpicture}
\caption{\textbf{Throughput (FPS):} FPS across varying batch sizes [1, 16, 32, 64, 128, 256, 512] for each model on NVIDIA L40s. Values averaged over 10,000 iterations per batch}
\label{fig:fps_vs_batchsize}
\end{figure}

\begin{figure}[htbp]
\centering
\begin{tikzpicture}
\begin{axis}[
  width=13cm,
  height=7cm,
  xlabel={Batch Size},
  ylabel={Latency (ms)},
  title={Latency vs Batch Size},
  legend style={
    font=\scriptsize,    
    draw=black,          
    fill=white,          
    fill opacity=0.9,    
    at={(0.98,0.98)},    
    anchor=north east,   
    legend cell align=left
  },
  ymajorgrids=true,
  grid style=solid,
  mark size=2.5pt
]

\addplot+[mark=*, thick] coordinates {(1,5.2)(16,0.5)(32,0.52)(64,0.55)(128,0.58)(256,0.6)(512,0.62)};
\addlegendentry{ConvNeXt-T~\cite{liu2022convnext}}

\addplot+[mark=*, thick] coordinates {(1,8.5)(16,0.6)(32,0.61)(64,0.63)(128,0.65)(256,0.66)(512,0.67)};
\addlegendentry{EfficientNetV2-S~\cite{tan2021efficientnetv2}}

\addplot+[mark=*, thick] coordinates {(1,3.2)(16,0.2)(32,0.22)(64,0.23)(128,0.24)(256,0.25)(512,0.26)};
\addlegendentry{MobileNetV3-L~\cite{howard2019}}

\addplot+[mark=*, thick] coordinates {(1,4.5)(16,0.45)(32,0.46)(64,0.47)(128,0.48)(256,0.49)(512,0.50)};
\addlegendentry{MobileViT v2 (S)~\cite{mehta2023mobilevitv2}}

\addplot+[mark=*, thick] coordinates {(1,5.0)(16,0.48)(32,0.49)(64,0.50)(128,0.51)(256,0.52)(512,0.53)};
\addlegendentry{MobileViT v2 (XS)~\cite{mehta2021mobilevit}}

\addplot+[mark=*, thick] coordinates {(1,3.5)(16,0.42)(32,0.43)(64,0.44)(128,0.45)(256,0.46)(512,0.47)};
\addlegendentry{RepVGG-A2~\cite{ding2021repvgg}}

\addplot+[mark=*, thick] coordinates {(1,4.3)(16,0.6)(32,0.61)(64,0.62)(128,0.63)(256,0.64)(512,0.65)};
\addlegendentry{TinyViT-21M~\cite{tinyvit2022}}

\end{axis}
\end{tikzpicture}
\caption{\textbf{Inference Latency:} Inference latency (ms) across batch sizes on an NVIDIA L40s GPU with 224×224 input resolution. Latency measured in PyTorch eager mode using AMP.}
\label{fig:latency_vs_batchsize}
\end{figure}

\subsection{Model Suitability for Real-Time Use}

Among all tested models, \textbf{MobileNetV3--L} and \textbf{RepVGG--A2} exhibited the greatest balance of high accuracy and low latency as shown in figure~\ref{fig:fps_vs_batchsize}~and~\ref{fig:latency_vs_batchsize}. At batch size 1, MobileNetV3--L maintained latency under 3.2~ms, while scaling to over 9800 FPS at batch size 512. These qualities make such models excellent for embedded deployment in mobile, surveillance, or industrial robotics applications.

\subsection{Reproducibility and Fair Comparison}

All experiments were conducted under controlled hardware and software environments, with the same input pipeline and augmentation strategies. The use of a class-balanced subset, though not equivalent to full ImageNet--1K, provides a reproducible and time-efficient benchmark for comparative hyperparameter evaluation. Full training logs and benchmarking scripts are included in the supplementary repository.

\section{Conclusion}\label{sec:conclusion}

This study isolates and quantifies hyperparameter sensitivity across multiple lightweight architectures, providing a reproducible benchmark for real-time deployment analysis. By training on a carefully curated, class-balanced 90,000--image subset of ImageNet--1K, we systematically assessed the influence of various hyperparameter strategies—-including learning rate schedules, data augmentation methods, and optimizer selection-—on both model accuracy and inference efficiency.

Our results demonstrate that hyperparameter tuning alone, without modifying the model architecture, can lead to substantial performance gains. Across the evaluated architectures—ConvNeXt--Tiny, EfficientNetV2 --S, MobileNetV3--Large, MobileViT v2 (S and XS), RepVGG-A2, and TinyViT-21M—optimized configurations resulted in Top-1 accuracy improvements ranging from 1.5\% to 3.5\% compared to baseline settings. These gains were achieved through the combined use of cosine learning rate scheduling, modern augmentations such as CutMix and RandAugment, and adaptive optimizers like AdamW and RAdam. Notably, TinyViT-21M achieved the highest Top-1 accuracy of 90.94\%, followed closely by EfficientNetV2-S and RepVGG-A2, with 88.50\% and 88.45\%, respectively.

In addition to classification accuracy, we benchmarked all models for real-time deployment feasibility using throughput (frames per second) and latency (milliseconds per image) metrics. These benchmarks were conducted on an NVIDIA L40s GPU using PyTorch’s eager mode with Automatic Mixed Precision (AMP). Batch sizes ranged from 1 to 512, reflecting both low-latency and high-throughput scenarios. Results showed that models such as MobileNetV3-L and RepVGG-A2 consistently delivered below 5 ms average latency and throughput exceeding 9,000 FPS at higher batch sizes, making them strong candidates for deployment on edge devices, mobile platforms, and embedded systems with strict latency requirements.

However, while our findings are promising, it is important to acknowledge the limitations of the study. The use of a 90K-image subset, although balanced and reproducible, does not fully replicate the complexity and diversity of the complete ImageNet-1K dataset. As such, absolute accuracy values should not be directly compared with those reported in original papers trained on the full dataset. Instead, our study focuses on relative improvements due to hyperparameter optimization and their implications for model efficiency and deployability.

To support reproducibility and future research, all training logs, evaluation scripts, YAML configuration files, and dataset sampling protocols have been included in the supplementary materials. These artifacts allow for the exact replication of experiments and facilitate transferability of our findings to other domains such as medical imaging, robotics, and low-power AI systems.

The research illustrates that even with confined datasets and conventional training hardware, considerable performance and efficiency benefits are feasible with systematic hyperparameter optimization. While our benchmarks were conducted on a high-end NVIDIA L40s GPU, future work should explore deployment feasibility on edge CPUs (e.g., Raspberry Pi 5, ARM Cortex-A series) or mobile NPUs (e.g., Qualcomm Hexagon, Apple Neural Engine). Such analysis would help assess the generalizability of optimized lightweight models across diverse hardware settings. Our subset-based assessment approach gives a useful benchmark for lightweight model selection and optimization. The insights discovered here are particularly relevant for real-time applications in limited contexts, and we hope they serve as a platform for subsequent explorations into efficient deep learning at the edge.

\bibliography{reference}

\bibliographystyle{plain}  

\subsection*{Acknowledgement}

This work was supported by the Variable Energy Cyclotron Centre (VECC), Department of Atomic Energy (DAE), Government of India (GoI), and the Homi Bhabha National Institute (HBNI), Department of Atomic Energy (DAE), Government of India (GoI), for providing comprehensive facilities and technical support essential to this research. The Department of Atomic Energy (DAE), Government of India (GoI), is also appreciated for financing the open-access publishing of this study. The authors appreciate the peer reviewers for their insightful remarks and constructive criticism, as well as the personnel of the VECC library for their essential support over the course of this research.

\subsection*{Author contributions statement}

VKR contributed to the conceptualization of the study, validation of results, and manuscript review and editing, and also played a role in refining the model architecture and ensuring experimental consistency. SM led the original draft preparation and was responsible for data curation, methodology design, model implementation, visualization, and detailed experimental investigation. TS provided critical technical guidance, assisting in model execution, computational setup, and troubleshooting, while also facilitating access to necessary resources and optimizing runtime configurations. HKP supervised the overall research process, ensuring adherence to academic standards and research ethics, and offered strategic input throughout the analytical workflow. AD offered expert mentorship and conceptual oversight, aligning the research direction with broader scientific objectives, and contributed significantly to the technical narrative and analytical refinement. All authors have reviewed and approved the final version of the manuscript and agree to be accountable for all aspects of the work.

\subsection*{Additional information}

\textbf{Accession codes}: Not applicable.\\
\textbf{Competing interests}: The authors declare no competing interests.\\
The corresponding author is responsible for ensuring that this statement is accurate and has been agreed upon by all co-authors.\\
All code, training logs, configuration files, and evaluation scripts used in this study are publicly available at~\url{https://github.com/VineetKumarRakesh/lcnn-opt}.

\subsection*{Declarations}

\subsubsection*{Funding}
This research was supported by the Variable Energy Cyclotron Centre (VECC), Department of Atomic Energy (DAE), Government of India (GoI), and the Homi Bhabha National Institute (HBNI), Department of Atomic Energy (DAE), Government of India (GoI).

\subsubsection*{Clinical Trial Number}
Clinical trial number: not applicable.

\subsubsection*{Ethics, Consent to Participate, and Consent to Publish}
Ethics, Consent to Participate, and Consent to Publish declarations: Not Applicable.

\subsubsection*{Conflict of Interest}
All authors assert that they own no financial or personal affiliations that may be seen as affecting the work provided in this study. Most of the authors, including the corresponding author, are either Technical or Scientific Officers at the Variable Energy Cyclotron Centre (VECC) or are associated with the Homi Bhabha National Institute (HBNI), both of which are under the Department of Atomic Energy (DAE), Government of India, which has rendered exclusive support for open-access publication. No conflicts of interest are acknowledged.

\newpage

\bmsection{Author Biographies}


\begin{table*}[h!]
\renewcommand{\arraystretch}{1.2}
\centering
\begin{tabular}{m{2.3cm} m{13cm}}

\raisebox{-0.5\height}{\includegraphics[width=2cm]{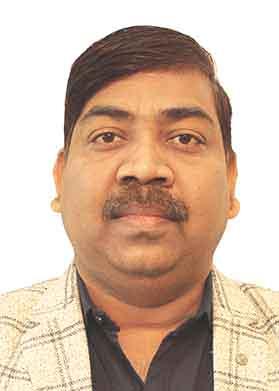}} &
\textbf{Vineet Kumar Rakesh} is a Technical Officer (Scientific Category) at the Variable Energy Cyclotron Centre (VECC), Department of Atomic Energy, India, with over 22 years of experience in software engineering, database systems, and artificial intelligence. His research focuses on talking head generation, lipreading, and ultra-low-bitrate video compression for real-time teleconferencing. He is pursuing a Ph.D. at Homi Bhabha National Institute, Mumbai. Mr. Rakesh has also contributed to office automation, OCR systems, and digital transformation projects at VECC. He is an Associate Member of the Institution of Engineers (India) and a recipient of the DAE Group Achievement Award. \\[5em]

\raisebox{-0.5\height}{\includegraphics[width=2cm]{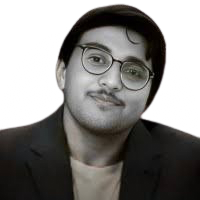}} &
\textbf{Soumya Mazumdar} is currently pursuing a dual degree: a B.Tech in Computer Science and Business Systems from Gargi Memorial Institute of Technology, and a B.S. in Data Science from the Indian Institute of Technology Madras. He has made significant contributions to interdisciplinary research, with over 25 publications in reputed journals and edited volumes by Elsevier, Springer, IEEE, Wiley, and CRC Press. His primary research interests include artificial intelligence, machine learning, 6G communications, healthcare technologies, and industrial automation. \\[5em]

\raisebox{-0.5\height}{\includegraphics[width=2cm]{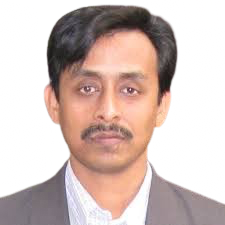}} &
\textbf{Dr. Tapas Samanta} is a senior scientist and Head of the Computer and Informatics Group at the Variable Energy Cyclotron Centre (VECC), Department of Atomic Energy, India. With a strong background in Electronics, Computer Science, and over two decades of experience, he leads initiatives in artificial intelligence, industrial automation, and control systems for particle accelerators. His work spans embedded systems, high-performance computing, and public technology outreach. Dr. Samanta also heads the Technology Transfer Division and Public Awareness Cell of VECC, promoting scientific innovation for societal benefit. \\[5em]

\raisebox{-0.5\height}{\includegraphics[width=2cm]{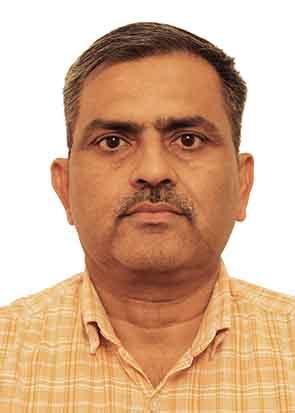}} &
\textbf{Dr. Hemendra Kumar Pandey} obtained his PhD from IIT Kharagpur.  He joined the Bhabha Atomic Research Centre (BARC) in Mumbai in 1999 and is presently employed as a Scientific Officer-G at the VEC Centre of the Department of Atomic Energy in Kolkata, where he has been engaged in the development of the Radioactive Ion Beam facility. This facility aims to provide accelerated beams of short-lived radioactive isotopes for research in nuclear physics, astrophysics, accelerator-based condensed matter physics, and related disciplines, as noted by a fellow of IETE.  He is an associate professor at the Homi Bhabha National Institute (HBNI), DAE Mumbai.  He has published over 102 research articles in national and international publications and conferences.  His research interests include RF systems for particle accelerators, beam diagnostics, high-power RF amplifier development, mixed-signal RF integrated circuit design, and radiation-hardened devices. \\[7em]

\raisebox{-0.5\height}{\includegraphics[width=2cm]{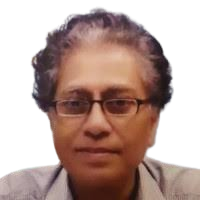}} &
\textbf{Dr. Amitabha Das} is the Director and Head of the School of Nuclear Studies and Application at Jadavpur University, Kolkata. He specializes in nuclear instrumentation, embedded systems, and reactor control systems. Dr. Das has led research on FPGA-based real-time data acquisition systems for high-resolution nuclear spectroscopy and has contributed to AI-driven applications such as lipreading and sign language recognition. He has supervised advanced research in nuclear reactor control, including LPV-based switching MRAC techniques. \\

\end{tabular}
\end{table*}

\end{document}